\documentclass[letterpaper, 10 pt, journal, twoside]{inc/ieeetran}
\usepackage[utf8]{inputenc}
\usepackage[T1]{fontenc}
\usepackage[english]{babel}

\def\BibTeX{{\rm B\kern-.05em{\sc i\kern-.025em b}\kern-.08em
    T\kern-.1667em\lower.7ex\hbox{E}\kern-.125emX}}

\usepackage{xurl}
\usepackage{color}
\usepackage{wrapfig}
\usepackage{textcomp}
\usepackage{makecell}
\newif\iffigs
\figstrue 

\usepackage[fleqn]{amsmath}
\usepackage{float}
\usepackage{setspace}
\usepackage{graphicx}
\usepackage{mathrsfs}
\usepackage{amssymb,amsfonts}
\usepackage{nicefrac}
\usepackage{algorithm}
\usepackage[noend]{algpseudocode}

\makeatletter
\newcommand\fs@spaceruled{\def\@fs@cfont{\bfseries}\let\@fs@capt\floatc@ruled
  \def\@fs@pre{\vspace{0.4\baselineskip}\hrule height.8pt depth0pt \kern2pt}%
  \def\@fs@post{\vspace{-0.4\baselineskip}\kern2pt\hrule\relax\vspace{-12pt}}%
  \def\@fs@mid{\kern2pt\hrule\kern2pt}%
  \let\@fs@iftopcapt\iftrue}
\makeatother

\usepackage{listings}
\lstnewenvironment{itemlisting}[1][]
 {%
  \mbox{}
  \vspace*{-\baselineskip}
  \lstset{
    xleftmargin=\leftmargin,
    linewidth=\linewidth,
    #1
  }%
 }
 {}
\usepackage{flushend}
\usepackage{multicol}

\usepackage{lipsum}

\usepackage[pdfa,colorlinks,bookmarksopen,bookmarksnumbered,allcolors=black,urlcolor=blue]{hyperref}
\usepackage[
	activate   = {true},
	protrusion = false,
	expansion  = true,
	kerning    = true,
	spacing    = true,
	tracking   = false,
	auto       = true,
	selected   = true,
	factor     = 1000,
	stretch    = 10,
	shrink     = 10,
]{microtype}

\usepackage[nameinlink,capitalise]{cleveref}
\crefname{line}{line}{lines}
\crefname{figure}{Fig.}{Figs.}
\Crefname{figure}{Fig.}{Figs.}
\crefname{equation}{Eq.}{Eqs.}
\Crefname{equation}{Eq.}{Eqs.}
\crefname{section}{Sec.}{Secs.}
\Crefname{section}{Sec.}{Secs.}
\crefname{definition}{Def.}{Defs.}
\Crefname{definition}{Def.}{Defs.}
\crefname{algorithm}{Alg.}{Algs.}
\Crefname{algorithm}{Alg.}{Algs.}
\crefname{assumption}{Asm.}{Asms.}
\Crefname{assumption}{Asm.}{Asms.}
\crefname{subassumption}{Asm.}{Asms.}
\Crefname{subassumption}{Asm.}{Asms.}
\Crefname{problem}{Problem}{Problems}
\crefname{problem}{Problem}{Problems}

\usepackage{cite}
\usepackage{amsmath,amssymb,amsfonts}
\usepackage{mathrsfs}
\usepackage{algorithm}
\usepackage{subfig}
\usepackage[export]{adjustbox}
\usepackage{algpseudocode}
\usepackage{multirow}
\usepackage{graphicx}
\usepackage{textcomp}
\usepackage{xcolor}
\usepackage[T1]{fontenc}
\usepackage{hyperref}

\usepackage{soul}
\usepackage{etoolbox}

\def\BibTeX{{\rm B\kern-.05em{\sc i\kern-.025em b}\kern-.08em
    T\kern-.1667em\lower.7ex\hbox{E}\kern-.125emX}}

\let\oldTH\TH
\renewcommand{\TH}{\textrm{\oldTH}}

\newcommand{\vect}[1]{\mathbf{#1}}

\newcommand{\EDIT}[2]{%
    {\color{blue}{%
    \ifstrempty{#1}%
        {%
            $\varnothing$ %
        }{%
            \st{#1} %
        }%
    $\rightarrow$ %
    \ifstrempty{#2}%
        {%
            $\varnothing$%
        }{%
            #2%
        }%
    }}%
}

\renewcommand{\EDIT}[2]{\textcolor{blue}{#2}}

\renewcommand{\EDIT}[2]{#2}
    
\begin{document}

\title{Robust Geospatial Coordination of Multi-Agent Communications Networks Under Attrition}

\markboth{IEEE Robotics and Automation Letters. Preprint Version. Accepted March, 2026}
{Kent \MakeLowercase{\textit{et al.}}: Robust Multi-Agent Networks Under Attrition} 

\author{\IEEEauthorblockN{Jonathan S. Kent$^{1,2,*}$, Eliana Stefani$^{3}$, Brian Plancher$^{4,5}$
\thanks{Manuscript received: November, 19, 2025; Revised February, 25, 2025; Accepted March, 23, 2026.}
\thanks{This paper was recommended for publication by Editor M. Ani Hsieh upon evaluation of the Associate Editor and Reviewers’ comments.} 
\thanks{$^{1}$JSK is with Advanced Technology Center, Lockheed Martin Space, USA and Fu Foundation School of Engineering and Applied Science, Columbia University, New York, USA {\tt\footnotesize jonathan.s.kent@lmco.com}}%
\thanks{$^{1}$ES is with Lockheed Martin AI Center, Lockheed Martin Corporation, USA {\tt\footnotesize eliana.stefani@lmco.com}}%
\thanks{$^{1}$BP is with Barnard College, Columbia University, New York, USA and  Dartmouth College, New Hampshire, USA {\tt\footnotesize plancher@dartmouth.edu}}%
\thanks{Digital Object Identifier (DOI): see top of this page.}%
\thanks{Authors thank Columbia's Junfeng Yang, CMU's Katia Sycara, Wennie Tabib, UIUC's Indranil Gupta, Darci Peoples, TU Braunschweig's Dominik M. Krupke, Georgia Tech's Hazel B. Sparks.}
}}

\maketitle

\begin{abstract}

\EDIT{Fast, efficient, robust communication during wildfire and other emergency responses is critical. One way to achieve this is by coordinating swarms of autonomous aerial vehicles carrying communications equipment to form an ad-hoc network connecting emergency response personnel to both each other and central command. However, operating in such extreme environments may lead to individual networking agents being damaged or rendered inoperable, which could bring down the network and interrupt communications.}
{Coordinating emergency responses in extreme environments, such as wildfires, requires resilient and high-bandwidth communication backbones. While autonomous aerial swarms can establish ad-hoc networks to provide this connectivity, the high risk of individual node attrition in these settings often leads to network fragmentation and mission-critical downtime.}
To overcome this challenge\EDIT{~and enable multi-agent UAV networking in difficult environments, this paper introduces and formalizes}{, we introduce and formalize} the problem of Robust Task Networking Under Attrition (RTNUA), which extends connectivity maintenance in multi-robot systems to explicitly address proactive redundancy and attrition recovery. We \EDIT{}{then} introduce Physics-Informed Robust Employment of Multi-Agent Networks ($\Phi$IREMAN), a topological algorithm leveraging physics-inspired potential fields to solve this problem. \EDIT{Through simulation across 25 problem configurations,}{In our evaluations,} $\Phi$IREMAN consistently outperforms \EDIT{the DCCRS baseline}{baselines}, and \EDIT{on large-scale problems, with up to 100 tasks and 500 drones, maintains $>99.9\%$ task uptime despite substantial attrition}{is able to maintain greater than $99.9\%$ task uptime despite substantial attrition in simulations with up to 100 tasks and 500 drones}, demonstrating both effectiveness and scalability.
\end{abstract}


\section{Introduction}

\EDIT{Consider a scenario involving the fighting of a forest or city fire, where emergency responders are distributed across multiple hotspots, each requiring constant communication with central command for effective coordination. The problem faced by these responders is not localized to a single point, but may be understood as a collection of tasks each requiring reliable communication (e.g., various outbreaks of flame, possibly moving over time). In such scenarios, communication to all agents at all task areas is essential to mission success, but it may be hampered by absent or destroyed cell infrastructure, or otherwise incur Size, Weight, and Power (SWaP) costs associated with satellite communications equipment~\cite{couceiro2019semfire,hu2022fault,arnold2018search}. One potential solution would be to deploy a swarm of networking drones to enable communication; each drone can move spatially, and when within range, communicate with other drones, the operator, and agents localized to each task. If the drones are spatially organized correctly, they will produce an ad-hoc network that allows agents at each task to communicate with their operator as well as with each other.}
{\IEEEPARstart{I}{n} large-scale emergency responses, such as urban or forest fires, responders are often distributed across multiple, shifting hotspots. Effective coordination in these scenarios relies on a constant flow of data between central command and the front lines. This problem is defined by a collection of dynamic tasks, each requiring a reliable communication backbone. However, infrastructure is frequently absent or compromised, and the high Size, Weight, and Power (SWaP) costs of satellite equipment often preclude its use~\cite{couceiro2019semfire,hu2022fault,arnold2018search}. A swarm of networking drones offers a viable alternative, autonomously organizing into an ad-hoc network to bridge this critical gap.}

\EDIT{However, due to the extreme environment, these network drones may undergo attrition, possibly disabled by heat, smoke, or bad actors~\cite{maresca2025react}. When a drone is attrited in this way, other drones must re-position themselves to re-establish lines of communication, but this still results in temporary disruption to the network. A robust system should therefore have the property that the drone network possesses enough slack and redundancy to ensure that limited drone attrition will not result in a disruption of connectivity to each task.}
{The environment in these scenarios is inherently hostile. Networking assets may undergo attrition, disabled by intense heat, smoke, or adversarial actions~\cite{maresca2025react}. When a node fails, traditional systems react by repositioning remaining drones to re-establish connectivity. However, this reactive approach inevitably incurs \emph{network downtime} during the transition, leaving responders isolated at critical moments. A truly robust system must anticipate this attrition, possessing enough inherent redundancy to ensure that connectivity remains uninterrupted even as the network self-heals.}

\EDIT{Similar challenges arise in}{Beyond firefighting, these challenges are pervasive in}
disaster response~\cite{queralta2020collaborative,ning2024multi}, search-and-rescue operations~\cite{agrawal2020next,maresca2025react}, and tactical battlefield engagements~\cite{yan2020distributed,ramachandran2024resilient}\EDIT{, where maintaining reliable communication networks is critical despite harsh conditions that may damage or destroy networking assets. The common thread across these scenarios is the need for a robust, self-healing network that can maintain connectivity even as individual nodes fail. 

Related problems in multi-robot systems have explored task allocation~\cite{shorinwa2023distributed,braquet2021greedy,prorok2016fast, liu2019submodular, gerkey2004formal, korsah2013comprehensive, zlot2006auction}, topology generation and maintenance~\cite{lee2023graph,tian2022kimera,luo2019minimum,MCCST,Lin-2021-128266}, spatial coordination~\cite{shi2020neural,riviere2020glas,luo2019minimum,MCCST,Lin-2021-128266}, and recovery from failures~\cite{colledanchise2015adaptive,senbaslar2023rlss,ramachandran2019resilience,goeckner2023attrition,dai2020multi,DCCRS}. However, \EDIT{much}{many} of these efforts do not consider the combined implications of these challenges.}
{. The common thread is the requirement for a self-healing architecture that maintains connectivity despite the loss of individual assets. 
While extensive research has explored task allocation~\cite{shorinwa2023distributed,braquet2021greedy,prorok2016fast, liu2019submodular, gerkey2004formal, korsah2013comprehensive, zlot2006auction}, topology maintenance~\cite{lee2023graph,tian2022kimera,luo2019minimum,MCCST,Lin-2021-128266}, spatial coordination~\cite{shi2020neural,riviere2020glas}, and failure recovery~\cite{colledanchise2015adaptive,senbaslar2023rlss,ramachandran2019resilience,goeckner2023attrition,dai2020multi,DCCRS} in isolation, few efforts have addressed the combined complexity of these challenges in a unified framework.}

\EDIT{This paper makes three primary contributions. First, we formalize the Robust Task Networking Under Attrition (RTNUA) problem combining task allocation, topology generation and maintenance, and attrition anticipation and recovery with quantitative uptime metrics. Second, we introduce the $\Phi$IREMAN algorithm\EDIT{}{,} achieving consistently higher task uptime than purely reactive approaches by proactively creating redundant network geometries via physics-inspired potential fields. Third, we provide comprehensive evaluation across 25 problem size and attrition rate configurations, demonstrating consistent improvements over \EDIT{the DCCRS baseline~\cite{DCCRS}.}{both MCCST~\cite{MCCST} and DCCRS~\cite{DCCRS} baselines.}}
{To overcome this challenge, this paper makes three primary contributions. First, we formalize the Robust Task-Networking Under Attrition (RTNUA) problem, integrating task allocation, topology maintenance, and attrition recovery with two new quantitative metrics. Second, we introduce $\Phi$IREMAN, a physics-inspired framework that proactively generates redundant network geometries by utilizing potential fields to maintain a resilient mesh. Third, we provide an extensive evaluation across 25 configurations of problem size and attrition rate, demonstrating that $\Phi$IREMAN consistently outperforms state-of-the-art baselines.}

\begin{table*}[t]
    \centering
    \small
    \begin{tabular}{c||c|c|c|c|c|c}
       Prior Work & \makecell{Task\\ Allocation} & \makecell{Topology\\ Generation} & \makecell{Topology\\ Maintenance} & \makecell{Spatial\\ Coordination} & \makecell{Attrition\\ Anticipation} & \makecell{Attrition\\ Recovery}\\
       \hline
       \hline
        \cite{hu2022fault,tian2022kimera} & \checkmark & \checkmark & \checkmark & \checkmark & & \checkmark \\
        \cite{shorinwa2023distributed,hudack2016multi} & \checkmark & & & & \checkmark & \\
        \cite{luo2019minimum} &  &  & \checkmark & \checkmark & \checkmark & \\
        \cite{Lin-2021-128266} & \checkmark &  & \checkmark & \checkmark &  & \\
        \cite{shi2020neural,riviere2020glas} & & & \checkmark & \checkmark & & \\
        \cite{colledanchise2015adaptive,halasz2007dynamic,prorok2016formalizing} & \checkmark & & & & & \checkmark \\
        \cite{senbaslar2023rlss} & \checkmark & & & \checkmark & & \\
        \cite{ramachandran2019resilience} & & \checkmark & \checkmark & \checkmark & & \checkmark \\
        \cite{goeckner2023attrition} & \checkmark & & \checkmark & & \EDIT{\checkmark}{} & \EDIT{}{\checkmark} \\
        \cite{dai2020multi} & \checkmark & & & & \checkmark & \checkmark \\
        \hline
        \cite{MCCST} (MCCST) & \checkmark &  & \checkmark & \checkmark &  & \\
        \cite{DCCRS} (DCCRS) & \checkmark & \checkmark & \checkmark & \checkmark & \checkmark & \checkmark \\
        \hline
        \hline
        Ours & \checkmark & \checkmark & \checkmark & \checkmark & \checkmark & \checkmark 
    \end{tabular}
    \caption{Feature sets and areas of concern for related work. Most works are not able to jointly handle task allocation and the generation and maintenance of network topologies in a geospatial setting, while anticipating and recovering from attrition\EDIT{. In later sections we demonstrate how our approach outperforms the DCCRS~\cite{DCCRS} baseline. Works are grouped according to features and concerns only, not on any basis of their operation.}{~based on features inherent in their design. Works are grouped according to features and concerns only, not on any basis of their operation. In later sections we demonstrate how our approach outperforms MCCST~\cite{MCCST} and DCCRS~\cite{DCCRS} baselines.}}
    \label{tab:capabilities}
    \vspace{-10pt}
\end{table*}


\section{Related Work}

\EDIT{Related problems in multi-robot systems have explored task allocation, topology generation and maintenance, spatial coordination, and recovery from failures. However, as noted above, much of this effort does not consider the combined implications of these challenges. \EDIT{We organize the discussion thematically to clarify how existing work addresses individual aspects, while the RTNUA problem requires their integration.}{}}
{Related problems in multi-robot systems span task allocation, topology maintenance, spatial coordination, and failure recovery. However, existing literature rarely addresses the intersection of these domains required by the RTNUA problem.}

\textbf{Multi-Robot Task Allocation.} 
\EDIT{Extensive work addresses multi-robot task allocation (MRTA)~\cite{shorinwa2023distributed, braquet2021greedy, prorok2016fast, liu2019submodular, gerkey2004formal, korsah2013comprehensive, zlot2006auction}. These approaches focus on assigning agents to tasks but do not address the critical networking behavior among agents, specifically determining geometric locations to ensure a connected network. For example, work such as~\cite{shorinwa2023distributed, hudack2016multi} optimizes assignments but treats communication as a constraint rather than a spatial coordination problem for co-design.}
{While extensive MRTA literature addresses agent-to-task assignment~\cite{shorinwa2023distributed, braquet2021greedy, prorok2016fast, liu2019submodular, gerkey2004formal, korsah2013comprehensive, zlot2006auction}, it largely treats communication as a secondary constraint rather than a spatial co-design problem~\cite{shorinwa2023distributed, hudack2016multi}, failing to actively position agents to guarantee network connectivity.}

\EDIT{\textbf{Network Topology Generation and Maintenance.} 
Research on topology generation and maintenance~\cite{lee2023graph, tian2022kimera, luo2019minimum, MCCST, Lin-2021-128266} ensures connectivity but typically assumes pre-allocated agents. Work on robust $k$-connectivity in swarm networks, such as~\cite{luo2019minimum}, assumes that each agent is already assigned to a given task and that the network maintains $k$-connectivity throughout. Crucially, these approaches provide no method for recovering connectivity if it is lost due to attrition. Similarly, while~\cite{MCCST} develops geometrically connected spanning trees among pre-tasked agents, it does not consider robustness or goal states. The algorithm in~\cite{Lin-2021-128266} jointly solves both the MRTA problem and the networking problem geometrically but again does not respond to or anticipate attrition.

\textbf{Spatial Coordination.} Approaches to spatial coordination~\cite{shi2020neural, riviere2020glas} achieve local interactions among robots but focus on collision avoidance or formation control rather than network topology optimization. While these methods demonstrate emergent behaviors from local rules, they do not explicitly optimize for communication network properties or task connectivity.}
{\textbf{Network Topology \& Spatial Coordination.}
Approaches to multi-agent swarm topology generation~\cite{lee2023graph, tian2022kimera, luo2019minimum, MCCST, Lin-2021-128266} ensure connectivity but typically assume pre-allocated agents and crucially lack mechanisms to recover from attrition. Even joint MRTA-geometry solvers~\cite{Lin-2021-128266} remain static. Similarly, spatial coordination methods~\cite{shi2020neural, riviere2020glas} prioritize local collision avoidance and formation control over optimizing global communication topologies for task execution.}

\textbf{Robustness to Failures.} 
\EDIT{Research focused on attrition within agent swarms typically centers on reacting to attrition once it occurs, rather than enabling robustness prior to attrition. Works such as~\cite{halasz2007dynamic, prorok2016formalizing, notomista2019optimal} primarily address task assignments rather than networking. While~\cite{ramachandran2019resilience} considers failures of particular resources on networked agents, it does not account for the complete loss of agents nor pre-configure the network to prevent temporary outages due to such attrition. The problem in~\cite{goeckner2023attrition} adapts to attrition of surveillance agents by modifying remaining agents' routes but does not plan ahead for attrition or address networking constraints. Although~\cite{hudack2016multi} plans routes to minimize capital loss due to attrition, it focuses on reducing attrition risk rather than maintaining network effectiveness despite assumed attrition. In~\cite{dai2020multi}, agent attrition results from direct combat, with fundamentally different dynamics.}
{Research on swarm attrition is predominantly reactive~\cite{halasz2007dynamic, prorok2016formalizing, notomista2019optimal}, focusing on task reassignment rather than network preservation. Other resilience strategies manage partial resource failures~\cite{ramachandran2019resilience}, adjust routes post-attrition~\cite{goeckner2023attrition}, minimize future attrition risk~\cite{hudack2016multi}, or model fundamentally different combat dynamics~\cite{dai2020multi}, but none proactively configure networks to absorb the complete loss of agents.}

\textbf{Positioning Our Contribution.} 
\EDIT{As shown in Table~\ref{tab:capabilities}, of these related works, only~\cite{DCCRS} treats every desired component of the RTNUA problem. The algorithm presented, denoted DCCRS in the absence of a name given in the manuscript, uses potential fields from attraction, repulsion, and velocity matching of neighbors, as well as specific modifications relating to density and distance-to-leader to maintain network coherence. However, DCCRS was designed only to retain network coherence under attrition, and while it implicitly has elements of recovery mechanisms, this was not a point of explicit design for DCCRS, nor was it treated in their experiments, which end as soon as decoherence occurs for the first time. Our approach considers all of these factors, and extends physics-based swarm modeling~\cite{pac2007control, merheb2016implementation, berlinger2021implicit, pimenta2013swarm} to organize drones into robust network geometries \EDIT{, and is demonstrated with metrics that capture both network maintenance and recovery capabilities}{. We evaluate using the $TU1$ and $TU2$ metrics introduced in Section~\ref{sec:rtnua}, which jointly capture network maintenance, recovery, and dispersion}.}
{As summarized in Table~\ref{tab:capabilities}, only DCCRS~\cite{DCCRS} addresses the full scope of the RTNUA problem. However, DCCRS was designed strictly to retain network coherence under attrition, lacks explicit recovery mechanisms, and its evaluation ceases at the first instance of decoherence. Our approach extends physics-based swarm modeling~\cite{pac2007control, merheb2016implementation, berlinger2021implicit, pimenta2013swarm} to actively organize drones into robust, self-healing network geometries. We evaluate these capabilities using the $TU1$ and $TU2$ metrics (introduced in Section~\ref{sec:rtnua}), which comprehensively capture network maintenance, recovery, and dispersion, including implementations of DCCRS and MCCST~\cite{MCCST} as baselines for comparison.}

\section{Robust Task Networking Under Attrition} \label{sec:rtnua}

To enable precise algorithm development and \EDIT{evaluation}{systematic benchmarking}, we formalize the RTNUA problem mathematically. The formalization captures three key requirements: drones must maintain network connectivity through spatial coordination, tasks require continuous connection to a base station, and random attrition degrades the network over time. We introduce two complementary metrics, $TU1$ measuring overall task uptime and $TU2$ discounting the \EDIT{exploration}{dispersion} phase, to evaluate algorithms comprehensively rather than simply measuring time-to-first-failure as in prior work.

\subsection{Problem Formulation}

\EDIT{We now formalize the Robust Task Networking Under Attrition (RTNUA) problem. A}
{An RTNUA problem is defined by a} set of $n$ drones $\mathcal{D} = \{D_1, D_2$\EDIT{$...$}{$, \dots$} $D_n\}$ \EDIT{is}{} distributed spatially about $\mathbb{R}^2$, with \EDIT{a controller existing at a base station $B$ located at the origin, with a communication radius $R$}{the base station controller $B$, with a communication radius $R$, at the origin}. Each drone $D_i$ is represented at time $t$ by the tuple $D_i(t) = \big(\vect{x}_i(t),$~\EDIT{$\lambda_i$}{$p_i$}$, r_i, v_i, e_i(t), a_i(t)\big)$, where:
\begin{itemize}
    \item $\vect{x}_i(t) \in \mathbb{R}^2$ \EDIT{gives its}{is the} position relative to the base station \EDIT{as a vector in $\mathbb{R}^2$}{}
    \item \EDIT{$\lambda_i \in \mathbb{R}_{>0}$ gives its attritional half-life}{$p_i(\delta t) \in [0, 1]$ is the probability of attrition in $\delta t$ time}
    \item $r_i \in \mathbb{R}_{>0}$ \EDIT{is its maximum radius of two-way communication}{is its maximum communication radius}
    \item $v_i \in \mathbb{R}_{>0}$ is its maximum speed
    \item $e_i(t) \in \{\texttt{TRUE, FALSE}\}$ \EDIT{gives its}{is the} current state of attrition, \texttt{FALSE} meaning it is alive and unattrited
    \item $a_i(t) \in \{\texttt{TRUE, FALSE}\}$ \EDIT{gives its}{is the} current activation state
\end{itemize}

\EDIT{Two drones}{In this model, two drones} $D_i, D_j$ can only communicate if $|\vect{x}_i - \vect{x}_j| \leq \min(r_i, r_j)$. For a given duration $\delta t$, each drone has the speed restriction $|\vect{x}_i(t+\delta t)-\vect{x}_i(t)| < v_i\delta t$. Drones start unattrited with $e_i(0) = \texttt{FALSE}$, and for duration $\delta t$, undergo attrition with probability \EDIT{$p_i(\delta t) = 1 - 0.5^{\delta t / \lambda i}$}{$p_i(\delta t)$}, such that $e_i(t+\delta t) = e_i(t) \lor \textrm{Bern}(p_i(\delta t))$. An attrited drone cannot be restored.\footnote{\EDIT{}{This model of attrition was selected for computational simplicity, and more complex or geographically localized attrition can be treated in future work.}}
A drone is considered active if it is both unattrited and network-connected (either directly or indirectly):
\begin{equation}
    a_i(t) = \neg e_i(t) \land \big(a_i^B(t) \lor a_i^d(t)\big),
\end{equation}
where, $a_i^B(t) = |\vect{x}_i(t)| \leq \min(R, r_i)$, and \\$a_i^d(t) = \big(\exists D_j \in \mathcal{D} \Big| a_j(t) \land |\vect{x}_i(t) - \vect{x}_j(t)| \leq \min(r_i, r_j)\big)$.

The system includes $m$ tasks $\mathcal{T} = \{T_1, T_2$\EDIT{$...$}{$, \dots$} $ T_m\}$. Practically speaking, each task would be an agent or set of agents requiring network connectivity, e.g. a group of firefighters, a member of the infantry, an unmanned ground vehicle, etc. Each task is represented as $T_i(t) = \big(\vect{y}_i(t), A_i(t)\big)$, with position $\vect{y}_i(t) \in \mathbb{R}^2$ and connection state $A_i(t) \in \{\texttt{TRUE, FALSE}\}$. A task is connected if at least one active drone is in range:
\begin{equation}
    A_i(t) = \exists D_j \in \mathcal{D} \Big| a_j(t) \land |\vect{y}_i(t) - \vect{x}_j(t)| < r_j.
\end{equation}

\EDIT{Denoting these as ``tasks'' is intended to stay \EDIT{in-line}{in line} with~\cite{luo2019minimum,MCCST}, and~\cite{Lin-2021-128266}, and \EDIT{are considered equivalent to the}{is equated to} ``leader drones'' from~\cite{DCCRS}.}
{Denoting these as ``tasks'' is intended to stay \EDIT{in-line}{in line} with~\cite{luo2019minimum,MCCST,Lin-2021-128266}, and they equate to ``leader drones'' from~\cite{DCCRS}.}

A controller $\mathscr{C}$ receives the set of networked drones $\mathcal{D}'(t) = \{D_i \in \mathcal{D} \big| a_i(t)\}$ and all tasks $\mathcal{T}$, providing motion instructions $\delta\vect{X}(t) = \{\delta \vect{x}_i(t)\}$ bounded by $|\delta \vect{x}_i(t)| \leq v_i$. For networked drones, $\vect{x}_i(t + \delta t) = \vect{x}_i(t) + \delta t \delta\vect{x}_i(t)$. For experimental purposes, we include a default behavior wherein unattrited but disconnected drones move directly toward the base station at constant speed until reconnected or destroyed, again leaving further development in this area to future work. 

\subsection{Objective Functions for Solution Benchmarking}

The objective functions for RTNUA are closely related to established metrics from multi-robot systems and network algorithm literature. Network uptime ratios, expressed as $U = (T_{\text{operational}}/\tau)$, have been \EDIT{extensively}{widely} used to quantify system reliability in surveillance and monitoring applications. Similarly, temporal connectivity measures such as connectivity persistence $CP = (1/\tau) \int_0^\tau a(t) dt$, where $a(t)$ indicates connectivity at time $t$, provide mathematical frameworks for evaluating time-varying network performance~\cite{sabattini2013decentralized, michael2009maintaining}.

Building on these foundations, we define the primary objective to maximize expected task uptime:

\begin{equation}
    TU1 = \mathbb{E}_{\textrm{scenario}}\Big[\int_{t=0}^{\tau} \frac{1}{m} \sum_{T_i \in \mathcal{T}} A_i(t) \partial t\Big]
    \label{eq:tu1}
\end{equation}

This formulation provides an appropriate communication uptime metric for the multi-task setting, \EDIT{where the proportion of time communication links remain active is generalized to the average connectivity across all tasks, where}{generalizing the proportion of time that communication links remain active to the average connectivity across all tasks, such that} a task is connected if it is within range of an active networked drone. The metric aligns with coverage quality metrics in wireless sensor networks that employ temporal coverage quality $\int_0^\tau C(t)dt$ where $C(t)$ represents the coverage function \cite{cardei2006energy, mini2012m}.

A secondary objective accounts for the speed of \EDIT{exploration}{dispersion} by only considering tasks after initial connection:

\begin{equation}
    TU2 = \mathbb{E}_{\textrm{scenario}}\Bigg[\int_{t=0}^{\tau} \frac{\sum_{T_i \in \mathcal{T}} A_i(t) \partial t}{\sum_{T_i \in \mathcal{T}} \exists t' \in [0, t) \Big| A_i(t')}\Bigg]
    \label{eq:tu2}
\end{equation}

This metric is related to the monitoring coverage ratios used in persistent surveillance literature, particularly in connectivity-constrained multi-robot persistent surveillance frameworks where robots must visit sensing locations periodically while maintaining network connectivity~\cite{scherer2020multi}. The normalization by discovered tasks reflects similar approaches in multi-UAV cooperative systems where performance metrics account for the proportion of targets under surveillance throughout the mission~\cite{wang2023spatio, su2023multi}. By providing this second metric, we can consider an algorithm's ability to maintain network uptime both in the context of and discounting its speed of \EDIT{exploration}{dispersion}.

The metric used in~\cite{DCCRS}, introducing DCCRS, was only the length of time until network decoherence first occurs. While this is meaningful for the problem of faultless network maintenance, it does not capture any information about the severity of failures, the magnitudes of their effects on users of the network, or how well the system might recover from failures afterwards. Our formulations are intended to capture these and other challenges of maintaining robust network connectivity despite ongoing attrition, while allowing for practical constraints like limited communication ranges and drone velocities.

\section{$\Phi$IREMAN Algorithm} \label{sec:phireman}

%

\EDIT{To provide a state-of-the-art solution to the RTNUA problem, we developed the Physics-Informed Robust Employment of Multi-Agent Networks, or $\Phi$IREMAN, algorithm as a computationally efficient baseline that leverages physics-inspired local interactions to achieve emergent robustness without requiring complex learning algorithms or extensive inter-agent communication.}
{The RTNUA formulation in Section~\ref{sec:rtnua} requires that a solution simultaneously allocates drones to tasks, generates and maintains a network topology connecting those tasks to the base station, and both anticipates and recovers from ongoing attrition, all while maximizing the time-averaged task connectivity captured by $TU1$ and $TU2$. Because these metrics penalize even brief connectivity interruptions across all tasks, an effective algorithm must not only react to attrition after it occurs, but also proactively establish redundant network geometries that can absorb individual failures without disconnecting any task. To address these requirements, we developed the Physics-Informed Robust Employment of Multi-Agent Networks, or $\Phi$IREMAN, algorithm, which leverages physics-inspired local interactions to achieve the desired robustness properties as emergent behaviors without requiring complex learning algorithms or extensive inter-agent communication.}

\EDIT{$\Phi$IREMAN is based upon leveraging the topological properties of efficient sphere packings in 2 dimensions, insofar as they spontaneously generate hexagonal meshes in low energy configurations. By combining a potential field that has been designed to encourage desired network geometries with synthetic attractive and repulsive forces akin to those observed in polar fluids at the molecular level, we can engineer an energy manifold for which low energy states represent the network occupying a designed geometry and exhibiting a hexagonal mesh pattern. By driving the combined network system towards those low energy states, we theorize and then observe regenerating network contiguity, robustness to attrition, exploration, and efficient spatial coverage all manifesting as emergent behaviors.}
{The core insight underlying $\Phi$IREMAN is that the topological properties of efficient sphere packings in two dimensions, i.e., that they spontaneously generate hexagonal meshes in low-energy configurations, naturally provide the path redundancy needed to sustain $TU1$ and $TU2$ under attrition, as any individual node in a hexagonal mesh may be removed without disconnecting its neighbors. By combining a task-space potential field, designed to guide drones toward an appropriate network geometry, with synthetic attractive and repulsive forces akin to those observed in polar fluids at the molecular level, we engineer an energy manifold whose low-energy states correspond to the drone network occupying the designed backbone geometry while generating and regenerating a hexagonal mesh pattern around it. Driving the system toward these low-energy states via gradient descent, we observe that task-directed dispersion, network contiguity, robustness to attrition, and recovery from connectivity loss all manifest as emergent behaviors, serving the $TU1$ and $TU2$ objectives without requiring explicit optimization against them.}

\subsection{The Computational Difficulties of Solving RTNUA}

Calculating the expected uptime of tasks under the RTNUA formulation presents significant computational challenges due to the fundamental intractability of percolation theory in continuum systems. Unlike discrete lattice percolation with some exact solutions for specific cases~\cite{kesten1982percolation}, continuum percolation systems lack general analytical solutions~\cite{balberg2009continuum, torquato2002random} and require computationally intensive Monte Carlo simulations~\cite{grimmett1999percolation}, making direct optimization approaches prohibitively expensive for real-time use. While decentralized multi-agent reinforcement learning (MARL) offers a general approach to distributed decision-making, robust MARL systems require extensive offline training~\cite{zhang2021multi}, significant inter-agent communication overhead scaling quadratically with agent count~\cite{foerster2016learning}, and substantial computational resources unsuitable for resource-constrained edge devices~\cite{chen2019deep}. 
\EDIT{These factors motivate the development of $\Phi$IREMAN as a computationally efficient baseline that leverages physics-inspired local interactions to achieve emergent robustness without requiring complex learning algorithms or extensive inter-agent communication, further differentiated from DCCRS~\cite{DCCRS} by removing velocity matching and leader following behaviors in favor of directly approximating a Steiner tree.}
{This motivates the development of $\Phi$IREMAN as a computationally efficient baseline that leverages physics-inspired local interactions to achieve emergent robustness without requiring complex learning algorithms or extensive inter-agent communication, and is further differentiated from DCCRS~\cite{DCCRS} by leveraging a Steiner tree approximation instead of velocity matching and leader following behaviors.}

\subsection{Task-Space Potential Field}



\begin{figure}[t]
    \centering
    \includegraphics[width=0.33\linewidth]{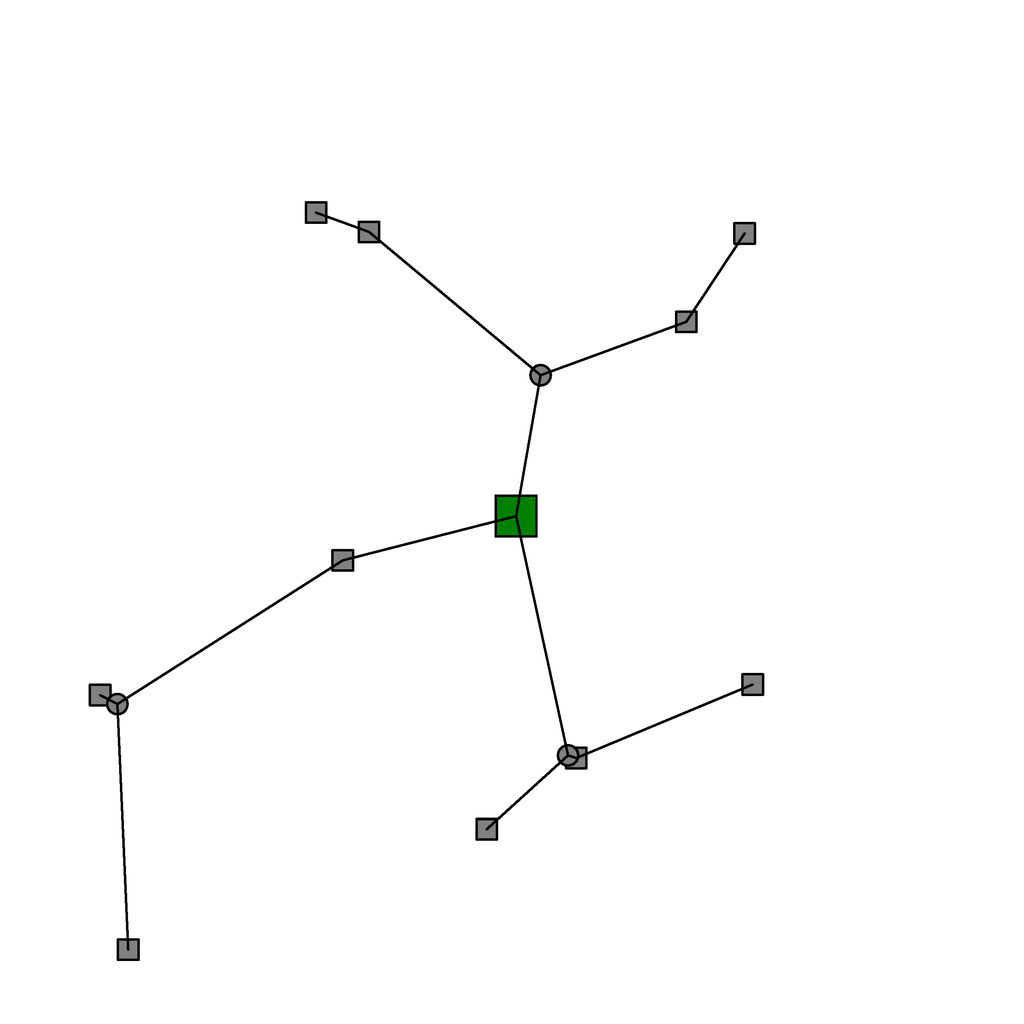}
    \includegraphics[width=0.54\linewidth]{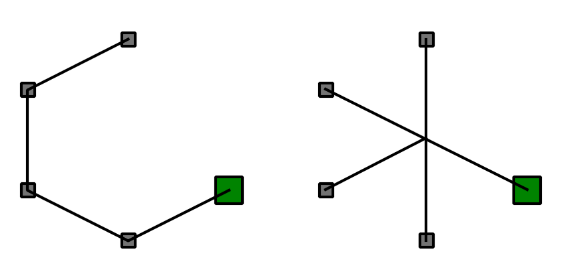}
    \caption{\EDIT{}{Left: }Semi-Steiner task tree example with $c_b = 0.3$, balancing total edge length ($l_{\Sigma e} = 19.86$) against sum of distances to base station ($l_{\Sigma b} = 39.53$). Steiner nodes (circles) enable more efficient network topologies. \EDIT{Left;}{Center:} hypothetical task tree graph using only sum of distances\EDIT{, right;}{. Right:} task tree graph including sum of distances to base station.}
    \vspace{-10pt}
    \label{fig:stein_example_horseshoe}
\end{figure}

\EDIT{We develop the geometry of the task space potential field by first constructing a graph that connects tasks to the base station.}{We develop the geometric structure underlying the task space potential field by first constructing a graph that connects tasks to the base station.} The potential at any location is then given by its distance to the nearest point on this graph. While a minimal spanning tree would minimize total edge length, it can produce geometries that are \EDIT{counter to robustness}{detrimental to robustness}, e.g. in the event that tasks form a horseshoe-shape with the base station located at one end, \EDIT{the minimal spanning tree would produce a geometry that would result in disconnection of the opposite end given an individual failure at any point.}{this would produce a geometry that would result in disconnection of the opposite end given an individual failure at any point (Figure \ref{fig:stein_example_horseshoe}, center and right).}

Instead, we jointly minimize the total graph length $l_{\Sigma e}$ and the sum of path distances from tasks to base station $l_{\Sigma b}$. The latter objective encourages more direct routes, reducing potential single points of failure. Additionally, since this is a geometric rather than purely topological problem, we greedily approximate a Steiner tree, inserting auxiliary Steiner nodes to reduce overall distance costs.
The weighting coefficient $c_b$ controls the trade-off between minimizing edge length and minimizing distances to base station. Higher values of $c_b$ produce more direct routes at the cost of increased total edge length. 
\EDIT{An example of a semi-Steiner task tree can be found in Figure~\ref{fig:stein_example_horseshoe}.}
{Figure~\ref{fig:stein_example_horseshoe}, left depicts a Semi-Steiner task tree.}

\subsection{Swarm Dynamics}

Rather than explicitly simulating fluid dynamics with momentum, we use gradient descent on an energy manifold to drive drones toward low-energy configurations. This replicates the desired fluid-like behaviors more efficiently in both computation and motion. The total potential energy for drone $D_i$ is given by \EDIT{}{the following, where $\TH \cdot h^\mathcal{T}(\vect{x}_i)$ represents the potential due to the task space potential field geometry, $\mathcal{M}$ represents the attraction between drones, and $P$ represents elastic repulsion between drones when closer than $q_i + q_j$:}
\begin{equation}
\begin{split}
  E_i^\Sigma &= E_i^{\TH} + E_i^\mathcal{M} + E_i^{P}, \quad \text{where}: \\
  &E_i^{\TH} = \TH \cdot h^\mathcal{T}(\vect{x}_i),  \quad
  E_i^\mathcal{M} = \mathcal{M} \cdot \sum_{D_j \in \mathcal{D} | i \neq j} \frac{-1}{|\vect{x}_j - \vect{x}_i|}, \\
  &E_i^{P} = \frac{P}{2} \cdot \sum_{D_j \in \mathcal{D}'} \big(q_i + q_j - |\vect{x}_i - \vect{x}_j|\big)^2.
\end{split}
\end{equation}
\EDIT{In this setup, we use $\TH \cdot h^\mathcal{T}(\vect{x}_i)$ to represent the potential due to the task space potential field geometry, $\mathcal{M}$ to represent the attraction between drones, and $P$ to represent elastic repulsion between drones when closer than $q_i + q_j$.}{}
We note that $\mathcal{D}'$ in the $P$ term enumerates drones within repulsion range and $q_i$ is nominally set to $r_i/2$ such that repulsion activates within communication range.

At each timestep, the controller $\mathscr{C}$ performs multiple steps of gradient descent on the total energy $\mathcal{E}(t, \delta \vect{X}(t)) = \sum_{D_i \in \mathcal{D}'(t)} E_i^\Sigma$ to generate motion instructions $\delta \vect{X}(t)$. Descent terminates when equilibrium is reached, maximum steps are taken, or any instruction would exceed drone speed limits. Motion instructions are then executed, with drones that have undergone attrition ceasing all movement. 

\EDIT{}{While gradient-based methods are susceptible to local minima, in the context of our swarm geometry, these minima represent stable, force-balanced configurations that fulfill the task-tree requirements. Because the energy landscape is dynamic, especially as drones move and undergo attrition, the system naturally explores the state space. Our analysis suggests that the current optimization approach generates sufficient topologies without the need for computationally expensive global search heuristics, as they effectively maximize the task uptime metrics as shown in Section~\ref{sec:eval}.}

\subsection{Network Maintenance}

As tasks move, Steiner node locations are continuously recalculated as the Fermat points of their three neighboring nodes. The complete task tree topology is periodically recomputed, with the new topology accepted if $l_{\Sigma}^{\textrm{new}} < \nu \cdot l_{\Sigma}^{\textrm{curr}}$, where $\nu \in (0, 1]$ is a threshold coefficient and $l_{\Sigma} = l_{\Sigma e} + c_b \cdot l_{\Sigma b}$.

\EDIT{\emph{Reaction to drone attrition requires no further extension.} The potential field naturally causes neighboring drones to flow into gaps while maintaining network connectivity. The redundancy inherent in the hexagonal packing means that a small number of losses can be absorbed without disrupting the network.}{\emph{Reaction to drone attrition requires no further extension.} The potential field naturally causes neighboring drones to flow into gaps left by attrited agents, preserving network connectivity. The redundancy inherent in the hexagonal packing means that a limited number of losses can be absorbed without disruption.}


\subsection{Asynchronicity and Message Passing}

\EDIT{To more closely resemble practical deployments, we introduce asynchronicity and message passing. Each drone maintains knowledge of its own position and broadcasts information to neighbors, received in the subsequent timestep. The $\Phi$IREMAN message passing scheme recursively forwards neighbor locations, retaining the most recent timestamp when receiving duplicate reports and discarding information that becomes too outdated or spatially distant, with thresholds determined via grid search. The Semi-Steiner tree is calculated centrally by the base station, which is assumed to have perfect, synchronous knowledge of task locations, and distributed throughout the network via packet forwarding, with updates following the same recency rule. Drones make kinematic decisions using only their local tree copy and locally known neighbor positions without accounting for information age. When the local task tree becomes too outdated, drones assume disconnection and navigate toward the known base station location as a recovery behavior, resuming normal operation upon re-establishing network contact.}
{To better reflect the constraints of practical deployments, we implement an asynchronous message-passing protocol to support a hybrid distributed-centralized architecture. Each agent maintains its own state and broadcasts it to its immediate neighbors, who receive the data in the subsequent timestep. While recursively forwarding information, agents retain only the most recent data based on timestamps and discard data that exceeds established temporal or spatial thresholds. While kinematic decisions and neighbor-coordination are computed locally, the high-level topological framework, that is the Semi-Steiner tree, is generated by a central base station with synchronous task-location data. This global topology is then propagated through the swarm using the same recursive forwarding logic. This hybrid approach allows drones to operate autonomously using local tree copies and navigate toward the base station to re-synchronize if network contact is lost.}

\section{Experimental Evaluation}\label{sec:eval}

\subsection{\EDIT{}{Baseline Performance}} \label{sec:res:baseline}
We evaluate $\Phi$IREMAN using a simulated environment with the following default configuration of both the underlying RTNUA problem and $\Phi$IREMAN algorithm, using parameters determined via a grid search:
\begin{itemize}
    \item 10 tasks uniformly distributed in $[-5, 5] \times [-5, 5] \subset \mathbb{R}^2$
    \item Base station at origin $(0,0)$
    \item 50 drones initialized in a radius-1 ring about the base station with Gaussian noise ($\sigma = 0.2$)
    \item Communication radius $r_i = 2$, repulsion radius $q_i = 1$
    \item Maximum drone speed $v_i = 0.15$, with a probability of attrition per-simulation step of 1\%
    \item Tasks random walk with Gaussian noise ($\sigma = 0.1$)
    \item Scenario length of 200 timesteps
    \item Task distance cost $c_b = 1.0$, topology threshold $\nu = 0.99$
    \item Potential field coefficients: attraction $\mathcal{M} = 0.05$, repulsion $P = 0.3$, task potential $\TH = 0.05$
\end{itemize}
$\Phi$IREMAN achieves task uptimes of $TU1 = 80.83\%$ and $TU2 = 85.34\%$. While this may appear modest compared to traditional network uptime metrics, it is achieved under substantial attrition with no \EDIT{hard-coded}{computationally explicit} recovery mechanism. 
\EDIT{}{Figure~\ref{fig:default_frames} shows the drone network evolution at $t = 0, 40, 80, 120, 160, 200$ timesteps.}
We note that this default setting is \EDIT{also a particularly challenging one, as on larger problem sizes, with up to 100 tasks,}{challenging, as on larger problem sizes that we explore shortly in our baseline comparisons,} \EDIT{exploration}{dispersion}-discounted task uptime, $TU2$, exceeds 99.9\%, even with substantial attrition. 

\subsection{Comparison Against \EDIT{DCCRS}{Baselines}}

\EDIT{To explore the relationship between algorithmic performance, problem size, and attrition rate, we provide comparisons between \EDIT{$\Phi$IREMAN and DCCRS~\cite{DCCRS}}{$\Phi$IREMAN, DCCRS~\cite{DCCRS}, and MCCST~\cite{MCCST}} over five different problem sizes and five different attrition rates, for a total of 25 configurations. Problem sizes range from extra small, ``XS'', to extra large, ``XL'', with values for the number of drones, number of tasks, and field size provided in Table~\ref{tab:psize}, as well as per-tick attrition rates of 0\%, 0.5\%, 1\%, 2\%, and 5\%.}
{To explore the relationship between algorithmic performance, problem size, and attrition rate, we provide comparisons between $\Phi$IREMAN and two baselines, DCCRS~\cite{DCCRS}, and MCCST~\cite{MCCST}, over five problem sizes and five attrition rates, for a total of 25 configurations. Problem size definitions are provided in Table~\ref{tab:psize}.}

\EDIT{For our implementation of DCCRS, hyperparameters with equivalents in $\Phi$IREMAN were carried over exactly, with all other hyperparameters determined via grid search for the default small, 1\% attrition rate setting, and held constant over other configurations.}{For our implementation of DCCRS, hyperparameters with equivalents in $\Phi$IREMAN were set to identical values, with all remaining hyperparameters determined via grid search for the default small, 1\% attrition rate setting, and held constant across other configurations.}

\EDIT{}{For our implementation of MCCST, to present the most faithful comparison possible to existing algorithms that include centralized components requiring synchronicity, drones were allowed to ``cheat'' by having perfect, instantaneous knowledge of the internal states of all other drones. A similar grid search approach was used to select hyperparameter values.} 

\begin{figure}[!t]
    \centering
    \subfloat[$TU1$ Scores]{\includegraphics[width=0.24\textwidth]{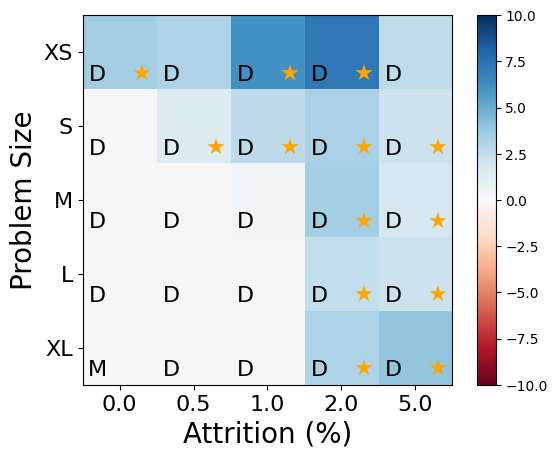}}
    \hfill
    \subfloat[$TU2$ Scores]{\includegraphics[width=0.24\textwidth]{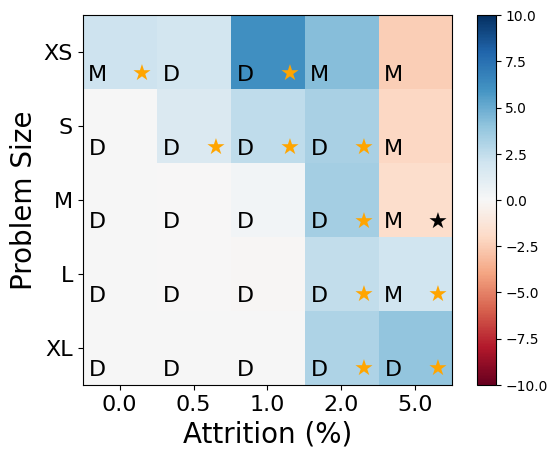}}
    \caption{Heatmap of differences in $TU1$ and $TU2$ scores between $\Phi$IREMAN\EDIT{and DCCRS}{, DCCRS, and MCCST} across problem configurations.
    \EDIT{Values shown are the result of subtracting the $TU1$ and $TU2$ scores of DCCRS from those of $\Phi$IREMAN for each configuration. Note that $\Phi$IREMAN generally outperforms DCCRS across all problems and has a maximum performance deficit of only 0.1 percentage points on $TU1$.}
    {Values shown are the difference of the mean $TU1$ and $TU2$ scores of $\Phi$IREMAN from the greater of DCCRS (D) or MCCST (M). Gold stars indicate that $\Phi$IREMAN outperformed both baselines to a degree statistically significant at the $p < 0.05$ level. Black stars indicate that one of the baselines similarly outperformed $\Phi$IREMAN. Overall, $\Phi$IREMAN statistically significantly outperforms baselines in 46\% of tests, is indistinguishable with scores over 95\% in 38\% of tests, and only statistically significantly lags baselines once.}}
    \label{fig:delta}
\end{figure}

We simulated 100 different scenarios for each configuration, with identical scenarios run against \EDIT{both $\Phi$IREMAN and DCCRS}{all three algorithms} by means of seeded random number generation, for a total of \EDIT{5,000}{7,500} simulations. Results are shown graphically as heatmaps in Figure \ref{fig:delta} and in Table~\ref{tab:delta}.

\begin{figure}[!t]
    \centering
    \includegraphics[width=0.8\linewidth]{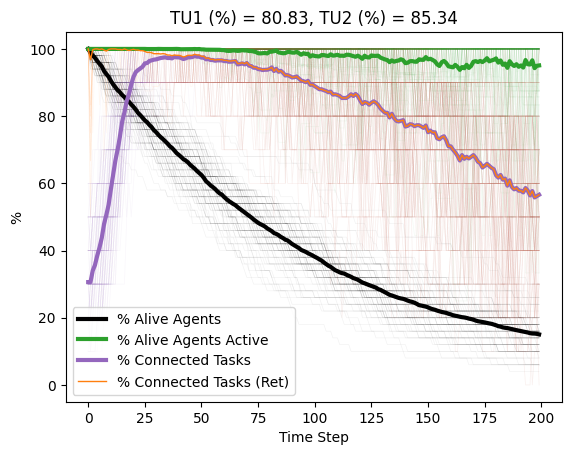}
    \caption{Performance over time of $\Phi$IREMAN under the default configuration. Thick lines give per-timestep averages over 100 simulations. ``Alive Agents'' shows how many drone agents remain alive; ``Alive Agents Active'' shows the portion of alive agents which are networked; ``Connected Tasks'' shows how many tasks are connected; and ``Connected Tasks (Ret)'' shows how many of the previously connected tasks are currently connected. Once all tasks are connected for the first time, these last two overlap. The integral of the former is $TU1$, and the integral of the latter is $TU2$.}
    \label{fig:default_graph}
    \vspace{-15pt}
\end{figure}
{
\captionsetup[subfloat]{captionskip=-5pt}
\begin{figure}[!t]
    \centering
    \subfloat[$t = 0$]{\includegraphics[width=0.15\textwidth]{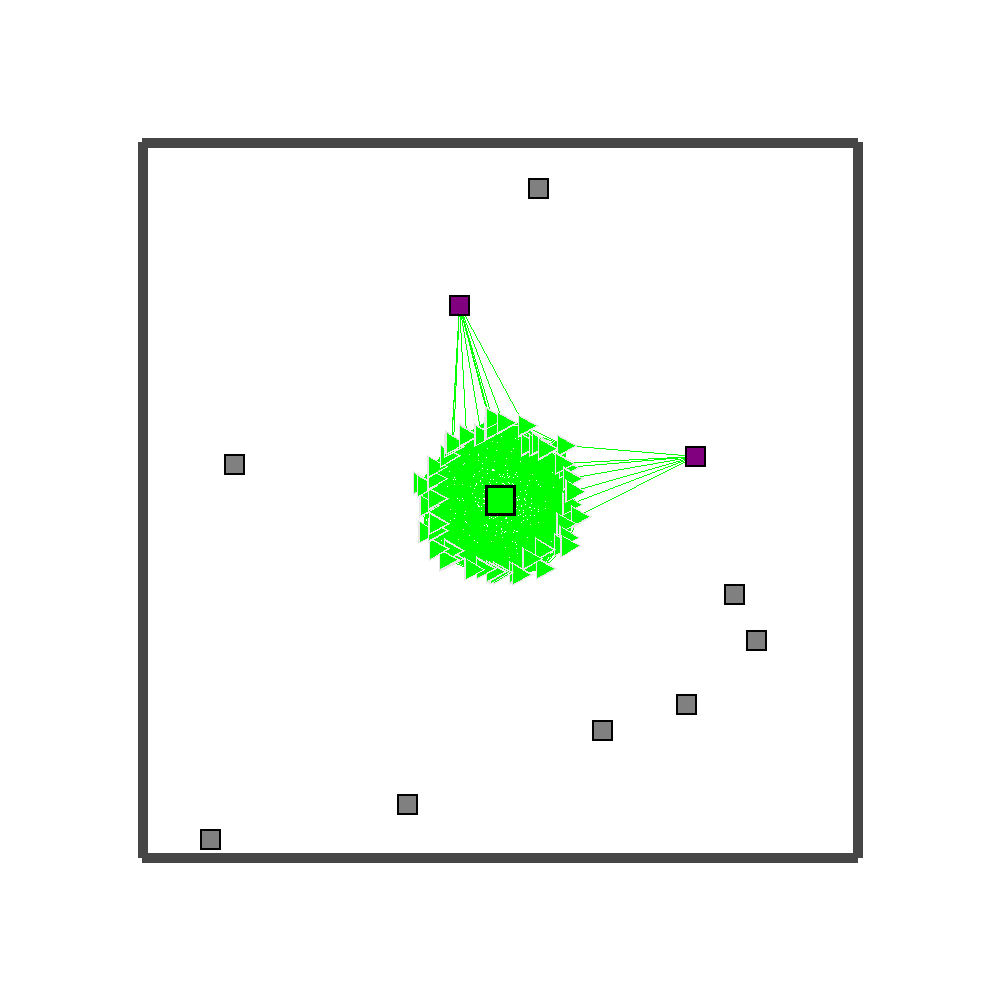}}
    \hfill
    \subfloat[$t = 40$]{\includegraphics[width=0.15\textwidth]{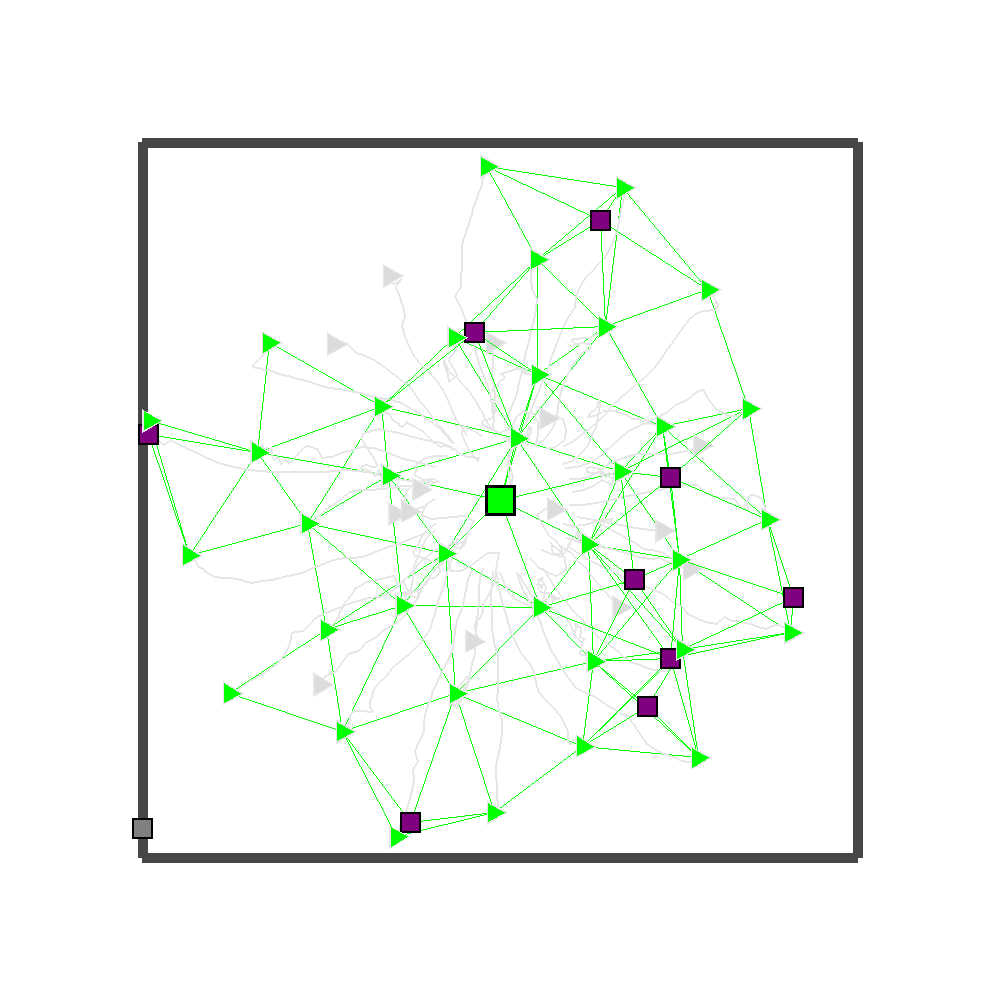}}
    \hfill
    \subfloat[$t = 80$]{\includegraphics[width=0.15\textwidth]{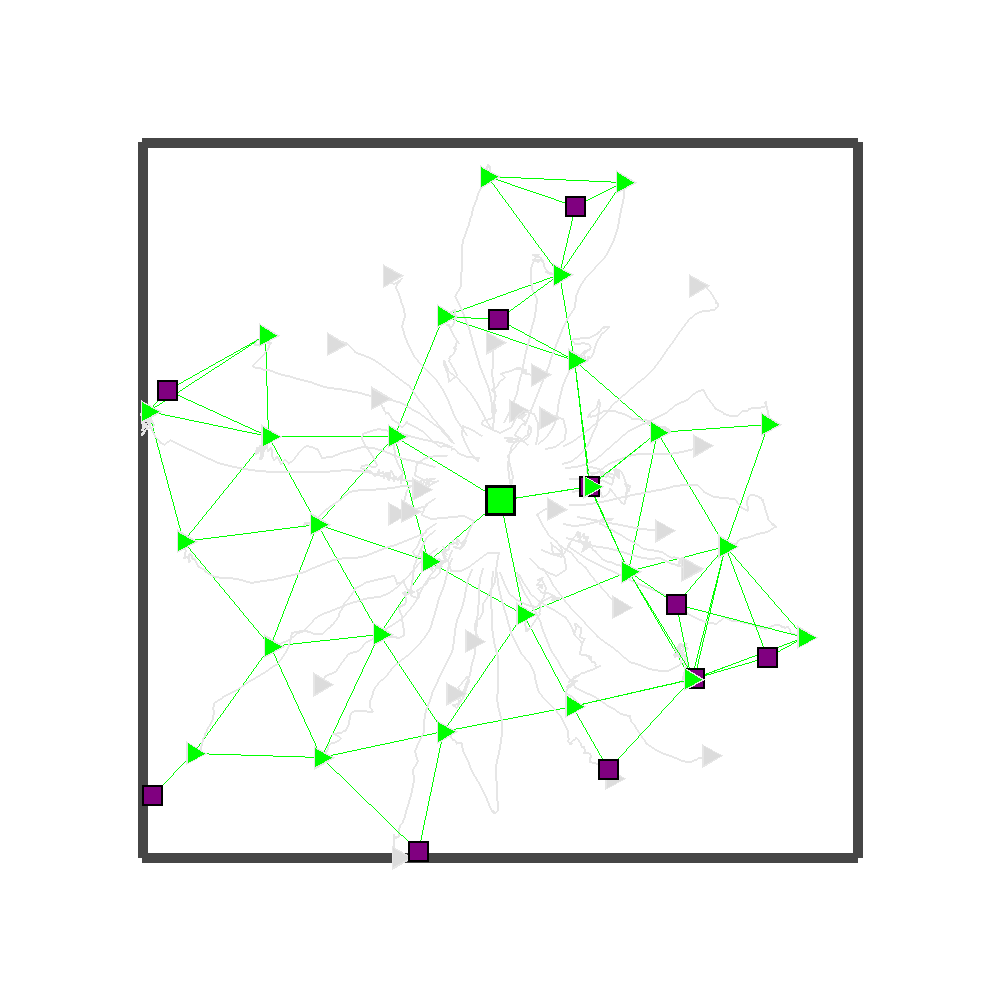}}\\
    \vspace{-10pt}
    \subfloat[$t = 120$]{\includegraphics[width=0.15\textwidth]{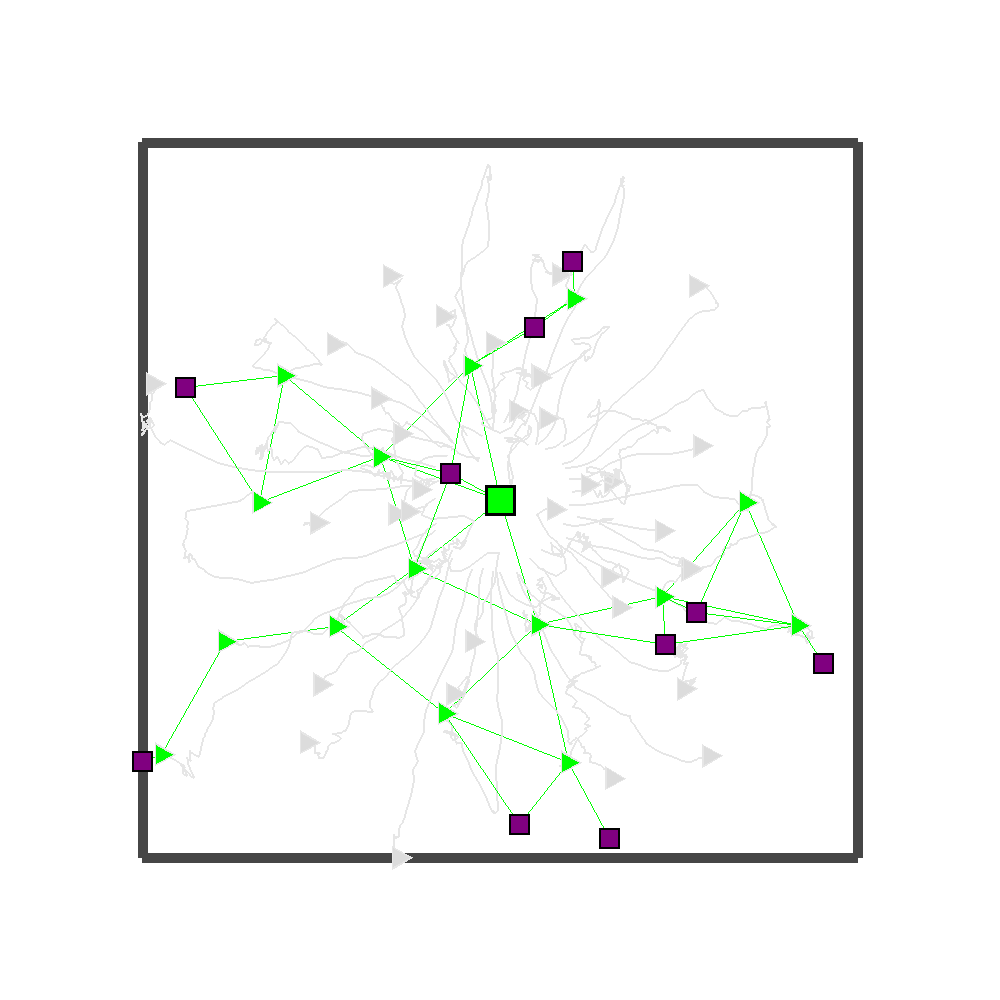}}
    \hfill
    \subfloat[$t = 160$]{\includegraphics[width=0.15\textwidth]{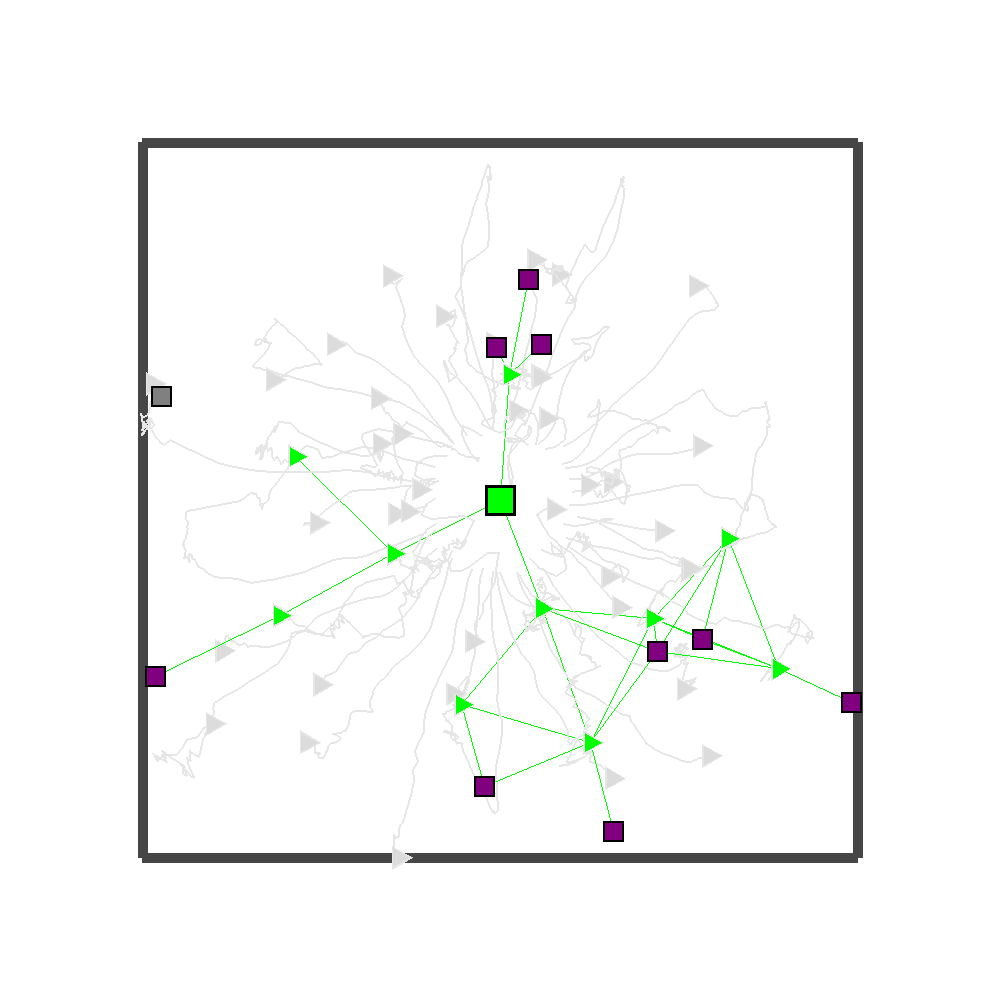}}
    \hfill
    \subfloat[$t = 200$]{\includegraphics[width=0.15\textwidth]{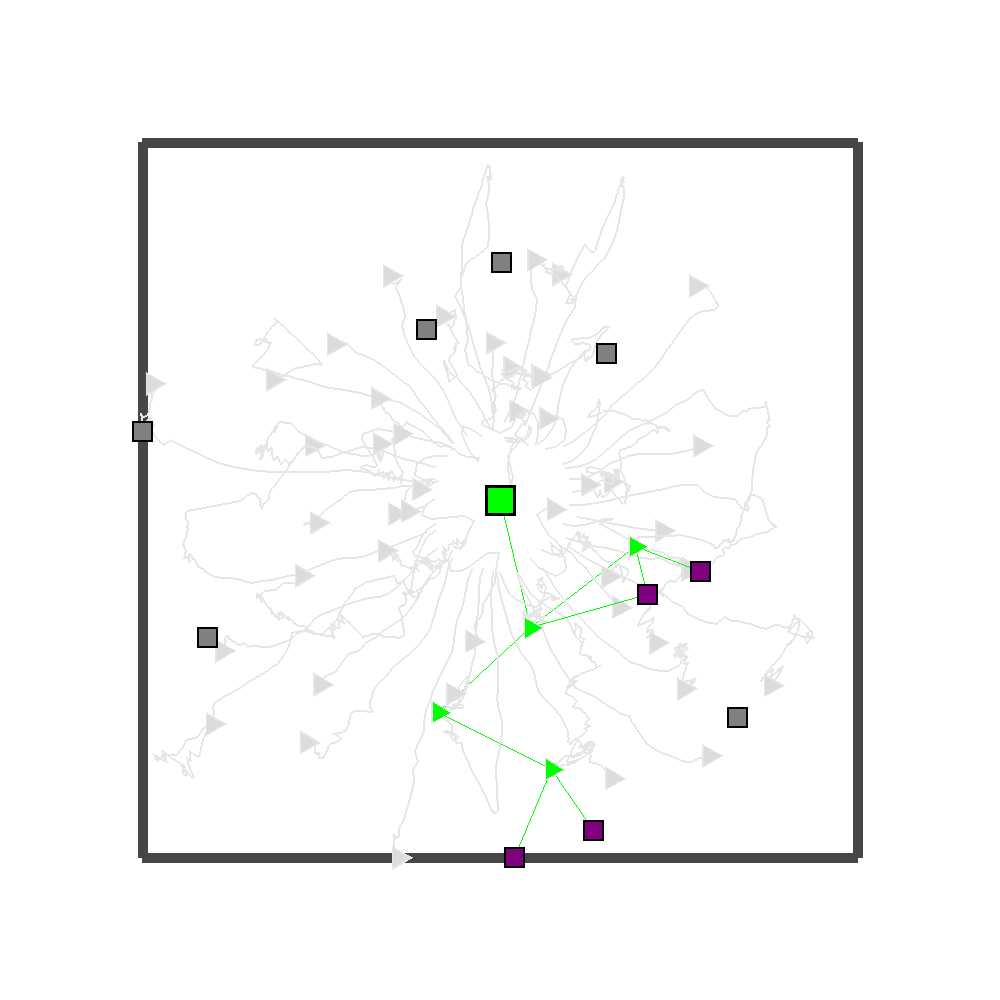}}
    \caption{Evolution of drone network under default configurations at $t = 0, 40, 80, 120, 160, 200$ timesteps. Alive drones (green triangles when active, red when disconnected) maintain connectivity between tasks (purple squares when networked, gray when disconnected) and base station (green square) despite ongoing attrition (gray triangles). The gray square is the field border. Green lines are network connections.}
    \label{fig:default_frames}
    \vspace{-15pt}
\end{figure}
}

\EDIT{These results show strong and consistent outperformance by $\Phi$IREMAN over \EDIT{DCCRS}{both DCCRS and MCCST}, with a maximum improvement of over 7 percentage points of $TU1$ in the extra small problem size with an attrition rate of 2\%, and a maximum performance deficit of only 1.1 percentage points of $TU1$ in the large problem size with an attrition rate of 1\%. These results demonstrate that the $\Phi$IREMAN algorithm performs better than the state-of-the-art.}
{These results show strong and consistent outperformance by $\Phi$IREMAN over both DCCRS and MCCST. Of the 25 configurations, $\Phi$IREMAN is statistically significantly better than both DCCRS and MCCST in 13 configurations for $TU1$ and in 10 configurations for $TU2$, and is statistically significantly outperformed only once.
$\Phi$IREMAN demonstrates a maximum improvement against both baselines of 7.18 percentage points of $TU1$ in the extra small problem size with an attrition rate of 2\%, being statistically significantly outperformed for the only time, by our ``cheating'' implementation of MCCST, with a performance deficit of 1.75 percentage points of $TU2$ in the medium problem size with an attrition rate of 5\%. Elsewhere, $\Phi$IREMAN outperforms MCCST by a maximum of 24.76 percentage points of $TU1$ in the small problem size with an attrition rate of 1\%.}

\subsection{Ablation Studies}

We additionally conduct ablation studies to analyze the impact of each algorithm and simulation component, with results presented in Table \ref{tab:param_results}. All ablations are done in the context of the default configurations presented in Section~\ref{sec:res:baseline}. These ablation studies suggest that $\Phi$IREMAN performs robustly across problem parameters, and that problem size and attrition rate have dominating effects on performance, as expected.


\begin{table}[!t]
    \centering
    \caption{Problem sizes for comparison studies.}
    \begin{tabular}{c|ccc}
         \textbf{Problem Size} & \textbf{Drones} & \textbf{Tasks} & \textbf{Field} \\
         \hline
         XS & 20  & 5   & $[-2, 2] \times [-2, 2]$ \\
         S  & 50  & 10  & $[-5, 5] \times [-5, 5]$\\
         M  & 100 & 20  & $[-10, 10] \times [-10, 10]$\\
         L  & 200 & 50  & $[-20, 20] \times [-20, 20]$\\
         XL & 500 & 100 & $[-50, 50] \times [-50, 50]$\\
    \end{tabular}
    \label{tab:psize}
    \vspace{-10pt}
\end{table}

\subsubsection{Task Graph Components}

Removing the task-space potential field entirely ($\TH = 0$) reduces performance to $TU1 = 75.76\%$, $TU2 = 80.79\%$. While drones form efficient hexagonal patterns with near-perfect inter-drone connectivity, task uptime suffers from lack of guidance toward task positions.
A similar yet slightly worse degradation to $TU1 = 75.21$\EDIT{}{$\%$}, $TU2 = 80.16$\EDIT{}{$\%$} occurs when the edges of the Semi-Steiner tree are removed from the task-space potential field, as drones tend to form clusters around tasks and do not receive stimulus to form a network between them.

\subsubsection{Environmental Factors}

%
Decreasing the drone speed to $v_i = 0.1$ reduces $TU1$ to $78.45\%$ while increasing the speed \EDIT{}{to} $v_i = 0.2$ increases $TU1$ to 82.06\%\EDIT{, while $TU2$ is largely unchanged from the default in either instance,}{. In both cases, $TU2$ remains largely unchanged from the default, }  \EDIT{, indicating}{and only the $TU1$ reduction at $v_i = 0.1$ is statistically significant. This indicates} that speed changes of this magnitude under these conditions primarily \EDIT{effects}{affect} \EDIT{exploration}{dispersion, and are of limited impact}.
Changing the standard deviation of the random walks of the tasks to $\sigma = 0$, rendering the tasks immobile, \EDIT{actually slightly decreases scores to $TU1 = 80.37\%$ and $TU2 = 85.10\%$, while}{has no statistically significant effect, and neither does} doubling task walk speed to $\sigma = 0.2$ \EDIT{reduces scores to $TU1 = 80.18\%$ and $TU2 = 84.48\%$}{}. \EDIT{The decrease with immobile tasks occurs because task motion provides natural exploration stimulus, causing drones to adjust positions and discover configuration improvements. With static tasks, drones converge to local energy minima that may not be globally optimal. This suggests future work should incorporate explicit exploration mechanisms for static task scenarios. The decrease with faster task motion demonstrates an increase in problem difficulty as expected.}{While the motion of tasks relative to the field and the existing drone network might be a source of dispersion noise that helps the system escape unstable equilibria in theory, we find it to be statistically negligible.}

\section{Conclusion and Future Work}

This paper introduces and formalizes the Robust Task Networking Under Attrition (RTNUA) problem and presents $\Phi$IREMAN (Physics-Informed Robust Employment of Multi-Agent Networks), which achieves robust networking through physics-inspired \EDIT{fluid dynamics
modeling to produce}{fluid-like potential field dynamics, producing} emergent behaviors that both anticipate and respond to attrition. Our experimental results demonstrate that $\Phi$IREMAN successfully produces redundant network topologies that maintain connectivity despite ongoing attrition, outperforming state-of-the-art baselines. Problem size and attrition rate emerge as primary constraints on achievable task uptime, with the task-tree potential field geometry proving essential for guiding drones toward effective network configurations.

\begin{table}[t]
    \centering
    \footnotesize
    \caption{
    \EDIT{Ablation study of algorithm components and environmental parameters on task uptime scores. Scores greater than 1 above the default configuration are shown in bold, scores less than 1 below the default configuration are italicized, and scores within $\pm 1$ of the default are shown in default text.}{Ablation study of algorithm components and environmental parameters on task uptime scores. Scores statistically significantly below the default configuration (at the $p < 0.05$ level) are italicized, while scores not statistically significantly different from the default are shown in default text. No scores statistically significantly above the default were found, confirming the validity of our results against ablation.}
    }
    \begin{tabular}{c|c|cc}
        & \textbf{Configuration} & $\mathbf{TU1}$ (\%) & $\mathbf{TU2}$ (\%)\\
        \hline
        & Default & 80.83 & 85.34 \\
        \hline
        \multirow{2}{*}{\shortstack[c]{Task\\ Graph}} & $\TH = 0$ & \textit{75.76 $\pm$ 1.84 } & \textit{80.79 $\pm$ 1.66 } \\
        & Nodes only & \textit{75.21 $\pm$ 1.67 } & \textit{80.16 $\pm$ 1.52 } \\
        \hline        
        \multirow{4}{*}{\shortstack[c]{Environmental\\ Factors}} & $v_i = 0.1$ & \textit{78.45 $\pm$ 1.69 } & 85.30 $\pm$ 1.44 \\
        & $v_i = 0.2$ & 82.06 $\pm$ 1.56 & 85.64 $\pm$ 1.47 \\
        & $\sigma = 0$ & 80.37 $\pm$ 1.81 & 85.11 $\pm$ 1.61 \\
        & $\sigma = 0.2$ & 80.18 $\pm$ 1.47 & 84.48 $\pm$ 1.39 \\
    \end{tabular}
    \label{tab:param_results}
    \vspace{-10pt}
\end{table}

\begin{table*}[!t]
    \centering
    \small
    \caption{
     \EDIT{$TU1$ and $TU2$ percentage scores for \EDIT{both $\Phi$IREMAN and DCCRS}{$\Phi$IREMAN, DCCRS, and MCCST} in a variety of problem configurations\EDIT{}{, including 95\% confidence intervals about the mean}. For a given problem configuration, treating $TU1$ and $TU2$ scores independently, \EDIT{where the difference between the performances of each algorithm is greater than 1, the better of the two results is presented in bold, otherwise in default text.}{wherever the best performance exceeds both competitors to a degree statistically significant at the $p < 0.05$ level, the best performance is presented in bold. All other scores are in default text}.  \EDIT{Note that $\Phi$IREMAN generally outperforms DCCRS across all problems and has a maximum performance deficit of only 0.1 percentage points on $TU1$}{In 13 configurations, $\Phi$IREMAN statistically significantly outperforms both DCCRS and MCCST on $TU1$, and on $TU2$ in 10 configurations. 19 configurations are statistically insignificant with $\Phi$IREMAN scoring over 95\% indicating regimes in which baseline performance was too high to demonstrate gains. $\Phi$IREMAN is only significantly outperformed once, by MCCST on $TU2$ in the medium problem size with the 5\% per-tick attrition rate, with the caveat that MCCST agents are permitted to ``cheat'' with instantaneous knowledge of the rest of the network}.}{
     We report $TU1$ and $TU2$ percentage scores for $\Phi$IREMAN, DCCRS, and MCCST across various configurations, including 95\% confidence intervals about the mean. For each configuration, $TU1$ and $TU2$ scores are analyzed independently. The top-performing algorithm is bolded only when its lead over all competitors is statistically significant at the $p < 0.05$ level according to a single-tailed Student’s $t$-test. $\Phi$IREMAN demonstrates meaningfully improved performance, statistically significantly outperforming both DCCRS and MCCST in 13 configurations for $TU1$ and 10 for $TU2$. Additionally, DCCRS achieves a score of greater than 95\% in 9 of the remaining 12 configurations for $TU1$, and 99\% in 9 of the remaining 15 configurations for $TU2$, greatly limiting the degree to which further improvements are possible. $\Phi$IREMAN was significantly outperformed only once, by MCCST on $TU2$ in the medium problem size with 5\% attrition, noting the caveat that our implementation of MCCST is permitted to ``cheat'' with all agents having instantaneous knowledge of the rest of the network.}
    }
    \begin{tabular}{cc|cc|cc|cc}
        \multicolumn{2}{c|}{\textbf{Configuration}} & \multicolumn{2}{c|}{\textbf{$\Phi$IREMAN}} & \multicolumn{2}{c}{\textbf{DCCRS}} & \multicolumn{2}{c}{\textbf{MCCST}}\\
        Size & Attrition (\%) & $\mathbf{TU1}$ & $\mathbf{TU2}$ & $\mathbf{TU1}$ & $\mathbf{TU2}$  & $\mathbf{TU1}$ & $\mathbf{TU2}$\\
        \hline
        \multirow{5}{*}{\rotatebox{90}{XS}} & 0.0 & \textbf{92.79 $\pm$ 0.98 } & \textbf{98.69 $\pm$ 0.48 } & 89.32 $\pm$ 1.65 & 96.46 $\pm$ 1.01 & 84.59 $\pm$ 1.75 & 96.58 $\pm$ 0.99 \\
        & 0.5 & 79.73 $\pm$ 2.62 & 86.97 $\pm$ 2.25 & 76.66 $\pm$ 2.98 & 85.11 $\pm$ 2.44 & 56.79 $\pm$ 3.84 & 70.04 $\pm$ 3.54 \\
        & 1.0 & \textbf{61.63 $\pm$ 3.14 } & \textbf{70.51 $\pm$ 3.05 } & 55.58 $\pm$ 3.55 & 64.37 $\pm$ 3.25 & 42.49 $\pm$ 3.24 & 57.12 $\pm$ 3.37 \\
        & 2.0 & \textbf{38.96 $\pm$ 3.43 } & 48.60 $\pm$ 3.48 & 31.78 $\pm$ 3.10 & 41.70 $\pm$ 3.09 & 29.10 $\pm$ 3.32 & 44.37 $\pm$ 3.67 \\
        & 5.0 & 21.54 $\pm$ 2.64 & 32.64 $\pm$ 3.08 & 19.00 $\pm$ 2.47 & 30.48 $\pm$ 3.09 & 18.17 $\pm$ 2.91 & 35.13 $\pm$ 4.25 \\
        \hline
        \multirow{5}{*}{\rotatebox{90}{S}} & 0.0 & 95.62 $\pm$ 0.26 & 99.95 $\pm$ 0.02 & 95.66 $\pm$ 0.27 & 99.91 $\pm$ 0.09 & 89.03 $\pm$ 0.77 & 98.66 $\pm$ 0.37 \\
        & 0.5 & \textbf{93.86 $\pm$ 0.57 } & \textbf{98.22 $\pm$ 0.42 } & 92.47 $\pm$ 0.63 & 96.78 $\pm$ 0.52 & 74.57 $\pm$ 2.30 & 85.49 $\pm$ 2.10 \\
        & 1.0 & \textbf{80.83 $\pm$ 1.63 } & \textbf{85.34 $\pm$ 1.51 } & 78.21 $\pm$ 1.84 & 82.78 $\pm$ 1.66 & 56.07 $\pm$ 2.84 & 68.18 $\pm$ 2.66 \\
        & 2.0 & \textbf{52.39 $\pm$ 2.28 } & \textbf{57.59 $\pm$ 2.05 } & 49.12 $\pm$ 2.27 & 54.32 $\pm$ 2.09 & 36.58 $\pm$ 2.36 & 49.48 $\pm$ 2.14 \\
        & 5.0 & \textbf{25.57 $\pm$ 1.69 } & 32.93 $\pm$ 1.57 & 23.39 $\pm$ 1.83 & 31.03 $\pm$ 1.91 & 20.42 $\pm$ 1.79 & 34.99 $\pm$ 2.03 \\
        \hline
        \multirow{5}{*}{\rotatebox{90}{M}} & 0.0 & 95.96 $\pm$ 0.17 & 99.98 $\pm$ 0.01 & 95.95 $\pm$ 0.17 & 99.97 $\pm$ 0.01 & 90.58 $\pm$ 0.56 & 98.67 $\pm$ 0.24 \\
        & 0.5 & 95.74 $\pm$ 0.19 & 99.80 $\pm$ 0.07 & 95.78 $\pm$ 0.18 & 99.83 $\pm$ 0.06 & 84.84 $\pm$ 1.16 & 93.66 $\pm$ 0.98 \\
        & 1.0 & 90.96 $\pm$ 0.72 & 95.05 $\pm$ 0.65 & 90.66 $\pm$ 0.79 & 94.74 $\pm$ 0.71 & 70.31 $\pm$ 1.76 & 79.86 $\pm$ 1.51 \\
        & 2.0 & \textbf{64.62 $\pm$ 1.56 } & \textbf{68.78 $\pm$ 1.48 } & 61.25 $\pm$ 1.62 & 65.34 $\pm$ 1.54 & 46.65 $\pm$ 1.70 & 57.43 $\pm$ 1.51 \\
        & 5.0 & \textbf{30.33 $\pm$ 1.32 } & 35.43 $\pm$ 1.19 & 28.62 $\pm$ 1.39 & 33.70 $\pm$ 1.25 & 24.37 $\pm$ 1.34 & \textbf{37.18 $\pm$ 1.36 } \\
        \hline
        \multirow{5}{*}{\rotatebox{90}{L}} & 0.0 & 96.12 $\pm$ 0.10 & 99.97 $\pm$ 0.01 & 96.13 $\pm$ 0.10 & 99.97 $\pm$ 0.01 & 93.35 $\pm$ 0.27 & 99.01 $\pm$ 0.13 \\
        & 0.5 & 96.09 $\pm$ 0.10 & 99.97 $\pm$ 0.01 & 96.10 $\pm$ 0.10 & 99.97 $\pm$ 0.01 & 92.00 $\pm$ 0.35 & 97.88 $\pm$ 0.23 \\
        & 1.0 & 95.41 $\pm$ 0.17 & 99.26 $\pm$ 0.12 & 95.51 $\pm$ 0.15 & 99.37 $\pm$ 0.11 & 85.29 $\pm$ 0.80 & 91.42 $\pm$ 0.70 \\
        & 2.0 & \textbf{75.86 $\pm$ 1.11 } & \textbf{79.75 $\pm$ 1.06 } & 73.43 $\pm$ 1.37 & 77.31 $\pm$ 1.34 & 57.90 $\pm$ 1.31 & 65.37 $\pm$ 1.26 \\
        & 5.0 & \textbf{36.15 $\pm$ 0.97 } & \textbf{40.27 $\pm$ 0.92 } & 34.02 $\pm$ 0.92 & 38.06 $\pm$ 0.89 & 28.55 $\pm$ 0.84 & 38.26 $\pm$ 0.89 \\
        \hline
        \multirow{5}{*}{\rotatebox{90}{XL}} & 0.0 & 96.21 $\pm$ 0.06 & 99.97 $\pm$ 0.00 & 96.22 $\pm$ 0.07 & 99.97 $\pm$ 0.00 & 96.29 $\pm$ 0.07 & 99.93 $\pm$ 0.02 \\
        & 0.5 & 96.22 $\pm$ 0.07 & 99.97 $\pm$ 0.00 & 96.21 $\pm$ 0.06 & 99.97 $\pm$ 0.00 & 96.13 $\pm$ 0.07 & 99.80 $\pm$ 0.04 \\
        & 1.0 & 96.20 $\pm$ 0.07 & 99.97 $\pm$ 0.00 & 96.19 $\pm$ 0.06 & 99.97 $\pm$ 0.01 & 95.26 $\pm$ 0.15 & 98.97 $\pm$ 0.13 \\
        & 2.0 & \textbf{88.63 $\pm$ 0.64 } & \textbf{92.41 $\pm$ 0.63 } & 85.56 $\pm$ 0.99 & 89.33 $\pm$ 0.98 & 77.80 $\pm$ 0.85 & 81.65 $\pm$ 0.84 \\
        & 5.0 & \textbf{43.79 $\pm$ 0.70 } & \textbf{47.60 $\pm$ 0.68 } & 39.82 $\pm$ 0.86 & 43.63 $\pm$ 0.83 & 37.85 $\pm$ 0.59 & 42.96 $\pm$ 0.57 \\
    \end{tabular}
    \label{tab:delta}
    \vspace{-10pt}
\end{table*}

Several promising directions remain for future work. To focus on formalizing RTNUA and establishing $\Phi$IREMAN's theoretical foundation, \EDIT{we restrict analysis to 2D and uniformly distributed, randomly walking tasks to provide}{the present work restricts analysis to two-dimensional environments with uniformly distributed, randomly walking tasks, providing} general results without application-specific assumptions. Future work may extend $\Phi$IREMAN to 3D or incorporate task motion prediction for increased deployed performance. 
\EDIT{Constraining drones within convex polyhedra fitted to task locations could improve efficiency, as drones positioned outside such regions cannot contribute to network redundancy more effectively than those inside. Adaptive parameter tuning mechanisms could balance competing objectives as network conditions evolve. Formal characterization of the relationship between environmental constraints and optimal algorithm parameters would enhance understanding.}{}
Future work could also explore large-scale multi-agent reinforcement learning approaches, which more directly optimize swarm performance against uptime~\cite{he2023robust, gupta2017cooperative, omidshafiei2017deep}. Finally, hardware demonstrations in field scenarios, paired with improved underlying local control~\cite{nguyen2024tinympc, song2023reaching}, would validate the algorithm's \EDIT{distributed nature and reliance on local interactions.}{real-world effectiveness in distributed settings.}

\bibliographystyle{inc/IEEEtran}
\bibliography{inc/main}

{\color{white}
{
    \centering
    \tiny{"A little song, a little dance, a little seltzer down your pants" -Tim}
}}
\color{black}

\end{document}